%% file: main.tex
\documentclass[11pt,a4paper]{article}

\usepackage[utf8]{inputenc}
\usepackage[T1]{fontenc}
\usepackage{lmodern}                    
\usepackage[margin=1in]{geometry}       
\usepackage{microtype}                  

\usepackage{amsmath,amssymb,amsfonts}
\usepackage{mathtools}                  

\usepackage{graphicx}
\usepackage{booktabs}                   
\usepackage{subcaption}                 
\usepackage{float}                      

\usepackage{algorithm}
\usepackage{algorithmic}

\usepackage{listings}
\usepackage{natbib}            
\usepackage{hyperref}
\hypersetup{
    colorlinks=true,
    linkcolor=blue,
    citecolor=blue,
    urlcolor=blue
}
\usepackage{cleveref}                   

\newcommand{\methodname}{Bilevel Autoresearch}
\newcommand{\valbpb}{\texttt{val\_bpb}}

\graphicspath{{figures/}}

\title{Bilevel Autoresearch: Meta-Autoresearching Itself}
\author{
    Yaonan Qu\thanks{Independent Researcher, EdwardOptimization@gmail.com}
    \and
    Meng Lu\thanks{Independent Researcher, menglu\_16@connect.hku.hk}
}
\date{}  

\begin{document}

\maketitle

\input{sections/abstract}

\input{sections/introduction}
\input{sections/related-work}
\input{sections/methods}
\input{sections/results}
\input{sections/discussion}
\input{sections/conclusion}


\bibliographystyle{plainnat}
\bibliography{references}


\end{document}

%% file: sections/abstract.tex
\begin{abstract}
If autoresearch is itself a form of research, then autoresearch can be applied
to research \emph{itself}.
We take this idea literally:
we use an autoresearch loop to optimize the autoresearch loop.

The autoresearch systems we build on---from Karpathy's single-track loop to
AutoResearchClaw's multi-batch extension and EvoScientist's persistent
memory---use search procedures designed by humans.
We ask whether an LLM can autonomously improve that procedure itself:
reading the loop, identifying bottlenecks, and generating new mechanisms that
change how future improvements are searched for.

We present \textbf{\methodname{}}, a bilevel framework where an outer loop
meta-optimizes the inner autoresearch loop by generating and injecting new
search mechanisms as Python code at runtime.
The inner loop optimizes the task; the outer loop optimizes how the inner
loop searches.
Both loops use the same LLM---no stronger model is needed at the meta level,
although the outer loop consumes additional inference and wall-clock budget.
This is a mechanism-level view of agentic self-improvement:
Python code is the carrier in our implementation, but skills, prompts,
workflows, scripts, evaluators, domain principles, world-model assumptions,
and memory schemas can also encode modifiable mechanisms that shape future
agent behavior.

On Karpathy's GPT pretraining benchmark, the meta-autoresearch outer loop
achieves a 5$\times$ improvement over the standard inner loop alone
($-0.045$ vs.\ $-0.009$ \valbpb{}), while parameter-level adjustment without
mechanism change yields no reliable gain.
The outer loop instantiates mechanisms from adjacent search domains, including
combinatorial optimization, multi-armed bandits, and design of
experiments---without human specification of the final mechanism design.
Trace analysis suggests that these mechanisms improve performance by breaking
the inner loop's deterministic search patterns, forcing exploration of
directions the LLM's priors systematically avoid.

The experiments demonstrate, on this benchmark, the first bilevel step:
an outer loop improves the search behavior of an inner loop.
The same mechanism-discovery principle suggests a path toward recursive
bootstrapping, where mechanisms discovered for the inner loop can be fed back
to improve the meta-level loop itself.
\end{abstract}

%% file: sections/introduction.tex
\section{Introduction}
\label{sec:introduction}

Large language models have demonstrated a striking capacity for self-directed
scientific iteration: given a task, an LLM can propose a change, execute an
experiment, observe the outcome, and decide whether to keep or discard the
change.
When repeated, this propose--execute--evaluate loop constitutes a form of
automated research~\citep{karpathy2026autoresearch}.
Instantiated for neural network hyperparameter search, we call this loop
\emph{autoresearch}.

At an abstract level, autoresearch is a closed-loop optimization process over
research actions.
A single LLM repeatedly proposing one intervention resembles sequential
single-point search.
Multi-batch autoresearch resembles batch or population search, and persistent
memory changes the state available to future proposals.
In the systems we build on, however, these search procedures are still fixed by
human designers.
\citet{karpathy2026autoresearch} introduced the single-track inner loop with a
keep/discard acceptance rule.
AutoResearchClaw~\citep{aiminglab2026claw} extended it with multi-batch parallel
search.
EvoScientist~\citep{evoscientist2026} added persistent experience memory across
runs.
These are useful mechanism choices, but the systems themselves do not discover
or replace the search procedure that produces future improvements.

This raises a natural question: \emph{can an outer loop perform that same
design step---reading code, identifying bottlenecks, writing new code---autonomously?}

\begin{figure}[!htbp]
  \centering
  \includegraphics[width=0.85\columnwidth]{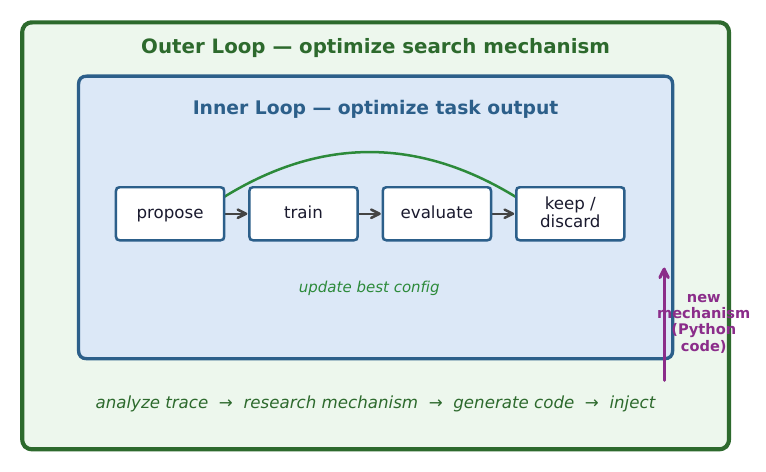}
  \caption{Bilevel autoresearch: the inner loop optimizes the task output;
  the outer loop optimizes the inner loop's search mechanism by generating
  and injecting new Python code at runtime.}
  \label{fig:concept}
\end{figure}

We provide an initial affirmative demonstration (\cref{fig:concept}).
We present \textbf{\methodname{}}, a bilevel framework with two
nested loops:
the \emph{inner loop} optimizes the task (proposing hyperparameter changes,
training, evaluating, keeping or discarding);
the \emph{outer loop} optimizes how the inner loop searches, by reading its
code, identifying bottlenecks, generating new Python mechanisms, and injecting
them at runtime.
Both loops use the same LLM, although the outer loop consumes additional
inference and wall-clock budget; any improvement comes from the bilevel
\emph{architecture}, not from a more capable model.

The central object is the mechanism, not the particular representation used in
this implementation.
We use Python code injection because code is explicit, executable, and easy to
validate before activation.
More generally, skills, prompts, workflows, scripts, evaluators, curricula,
domain principles, world-model assumptions, and memory schemas can also serve
as carriers of mechanisms that alter future agent behavior.
\methodname{} therefore targets mechanism-level self-improvement: improving how
an agentic loop generates, evaluates, selects, and revises future changes, not
only improving a one-off artifact.

This framing also suggests a recursive bootstrapping path.
If Level~2 discovers mechanisms that improve Level~1, the same discovery
principle can in principle be applied to the meta-level loop itself.
This paper empirically studies the first bilevel step---an outer loop improving
an inner loop---and leaves full recursive self-application to future work.

We evaluate the framework on Karpathy's GPT pretraining benchmark
with a controlled four-group ablation (\cref{sec:methods}).

\paragraph{Contributions.}
\begin{enumerate}
  \item We formulate autoresearch as a closed-loop search process and introduce
        a bilevel framework in which an outer loop optimizes the inner loop's
        search/update mechanism.
  \item We implement the outer level via a 4-round LLM dialogue that generates
        and injects new search mechanisms as Python code at runtime.
  \item A controlled four-group ablation shows that mechanism research
        (Level~2) produces a 5$\times$ improvement over the inner loop alone
        ($-0.045 \pm 0.030$ vs.\ $-0.009 \pm 0.002$), while parameter-level
        adjustment (Level~1.5) yields no reliable gain.
  \item Trace analysis suggests why: the generated mechanisms (Tabu Search,
        Bandit, Orthogonal Exploration) force exploration of directions the
        LLM's default search path systematically avoids.
  \item We clarify the scope of the claim: code is one carrier of modifiable
        mechanisms, and recursive self-application is a natural extension
        rather than an empirical result established in this paper.
\end{enumerate}

%% file: sections/related-work.tex
\section{Related Work}
\label{sec:related}

\subsection{Autoresearch and LLM-Driven Optimization}

\citet{karpathy2026autoresearch} introduced the paradigmatic autoresearch loop
for neural network hyperparameter search: an LLM reads a training script,
proposes a configuration change, executes training for a fixed budget, measures
validation loss, and accepts or rejects the change.
Iterated, this constitutes a form of LLM-guided hill climbing in configuration
space, where the LLM's world knowledge serves as an implicit prior over
promising changes and training outcomes provide gradient-free feedback.

AutoResearchClaw~\citep{aiminglab2026claw} extends this framework with
multi-batch parallelism: several candidate configurations are evaluated
simultaneously, and the best is promoted.
This increases the effective branching factor of search without altering the
underlying acceptance mechanism.

EvoScientist~\citep{evoscientist2026} introduces persistent experience memory:
lessons from prior runs are summarized and injected into future proposals,
enabling cross-run learning.
Both of these enhancements were designed by human researchers who inspected the
prior system's code and identified architectural gaps.
In these systems, the structural decisions---when to accept, how to propose,
what state to maintain---are made by human designers, not by the system itself.

\subsection{Bilevel Optimization}

Bilevel optimization~\citep{colson2007overview,sinha2018review} studies
problems of the form
$\min_{\phi}\, F\!\left(\phi,\, \theta^{*}(\phi)\right)$
subject to $\theta^{*}(\phi) \in \arg\min_{\theta}\, f(\theta, \phi)$,
where an upper-level objective $F$ depends on the optimal solution of a
lower-level problem parameterized by $\phi$.
Applications include meta-learning~\citep{franceschi2018bilevel},
neural architecture search~\citep{liu2018darts}, and hyperparameter
optimization~\citep{feurer2019hyperparameter}.
In our setting the upper level optimizes the search \emph{mechanism} $\phi$
(the runner code) and the lower level optimizes the task performance $\theta$
(the training configuration).
The key departure from classical bilevel optimization is that $\phi$ is a
program---a discrete artifact produced by code generation---rather than a
real-valued parameter vector.
We use the bilevel terminology in a structural and algorithmic sense:
the present system does not solve the classical nested optimization problem
exactly, but implements a finite-budget, non-differentiable analogue in which
the upper level modifies the lower level's search/update mechanism.

\subsection{LLM-Based Code Generation for Research}

AlphaCode~\citep{li2022competition} and Codex~\citep{chen2021evaluating}
demonstrated that LLMs can write functionally correct programs from natural
language specifications.
FunSearch~\citep{romera2024mathematical} extended this to scientific
discovery, using an LLM to iteratively generate and evaluate mathematical
programs, finding new results in combinatorics.
Most directly related is the line of work on LLM-driven algorithm
design~\citep{liu2024evolution,lehman2023evolution}, in which LLMs propose
novel algorithmic variants that are then evaluated on benchmark tasks.
Our Level~2 agent applies the same code generation capacity to a different
target: rather than generating task-level programs, it generates
\emph{search mechanism code} that is injected into the inner loop at runtime.

\subsection{Meta-Learning and Algorithm Configuration}

Meta-learning~\citep{hospedales2021meta} trains models to learn efficiently
from few examples by optimizing across a distribution of tasks.
Algorithm configuration~\citep{hutter2011smac} and algorithm
selection~\citep{rice1976algorithm} choose among candidate algorithms or
parameter settings for a given problem instance.
Portfolio methods~\citep{xu2008satzilla} maintain a library of algorithms and
select among them.
\methodname{} operates in a similar spirit---the outer loop selects or
generates a search mechanism---but uses LLM code generation rather than
gradient-based meta-optimization or a fixed portfolio.

\subsection{Artifact-Level and Mechanism-Level Self-Improvement}

Many agent self-improvement systems modify external artifacts that shape future
behavior, such as prompts, reusable skills, workflows, memory, harness state,
or code snippets.
These artifacts matter because they influence what the agent will do next.
However, improving an artifact is distinct from improving the mechanism that
generates, evaluates, revises, and deploys future artifacts.
Prompt and pipeline optimization methods such as ADOPT~\citep{zhao2025adopt}
treat language artifacts as optimizable parameters inside a fixed multi-step
LLM pipeline.
Principle-evolution methods such as PiEvo~\citep{pu2026pievo} target a
different carrier: they treat scientific discovery as Bayesian optimization
over an expanding principle space, allowing agents to refine the scientific
priors and worldview that guide hypothesis generation.
\methodname{} targets this mechanism-level object directly.
The current implementation represents mechanisms as injectable Python code, but
the same formulation can apply to other carriers when they change the
proposal, evaluation, selection, or update behavior of an agentic loop.

\paragraph{Concurrent and subsequent work.}
Since the initial March 24, 2026 arXiv version of this paper, several related
systems have explored skill-, artifact-, harness-, and process-level
self-improvement.
Meta-Harness~\citep{lee2026metaharness} searches over harness code for LLM
applications, while Continual Harness~\citep{karten2026continualharness}
adapts prompts, sub-agents, skills, and memory online from trajectories.
GEAR~\citep{jeddi2026gear} replaces single-path autoresearch with
population-based search over multiple research states and studies a variant
whose controller can evolve during the run.
Discovering Cooperative Pipelines~\citep{gallego2026cooperativepipelines}
studies two-level autoresearch for sequential social dilemmas, where an outer
coding agent edits prompts, feedback functions, helper libraries, and iteration
logic in an inner policy-synthesis pipeline.
Skill-focused work optimizes reusable skill artifacts: SkillEvolver frames
skill learning as a meta-skill~\citep{zhang2026skillevolver}, SkillOpt trains a
skill document as external agent state~\citep{yang2026skillopt}, and
\citet{huang2026bilevelskills} formulate skill structure and content
optimization as a bilevel problem.
AEvo~\citep{zhang2026aevo} is especially close in spirit: a meta-agent edits
the procedure or agent context that controls future evolution rather than
directly proposing the next candidate.
We view these works as complementary subsequent examples of external agentic
artifacts as useful intervention surfaces.
In our framing, however, a harness is one mechanism carrier among others:
skills, code, prompts, workflows, evaluators, scientific principles,
world-model assumptions, and memory schemas can all encode the proposal,
evaluation, selection, or update logic that shapes future improvements.

\subsection{Position of This Work}

The key distinction between \methodname{} and the autoresearch systems above is
the \emph{target} of the outer loop.
Level~1.5 is closest to existing outer loops (curriculum schedulers, adaptive
configuration) in that it adjusts parameters of the existing mechanism.
Level~2 is categorically different: it generates code that modifies or
replaces structural components of the search mechanism.
We are not aware of earlier autoresearch systems in which an autonomous outer
loop writes and injects code that modifies the structural logic of the inner
autoresearch loop at runtime, using the same model that runs the inner loop.
Many subsequent skill, harness, and workflow systems can be viewed through this
lens as related carrier-level instantiations: they often optimize useful
artifacts or agent scaffolding, while \methodname{} frames such artifacts as
possible representations of a deeper search/update mechanism.

%% file: sections/methods.tex
\section{Methods}
\label{sec:methods}

\subsection{Framework Overview}

\begin{figure}[!htbp]
  \centering
  \includegraphics[width=\columnwidth]{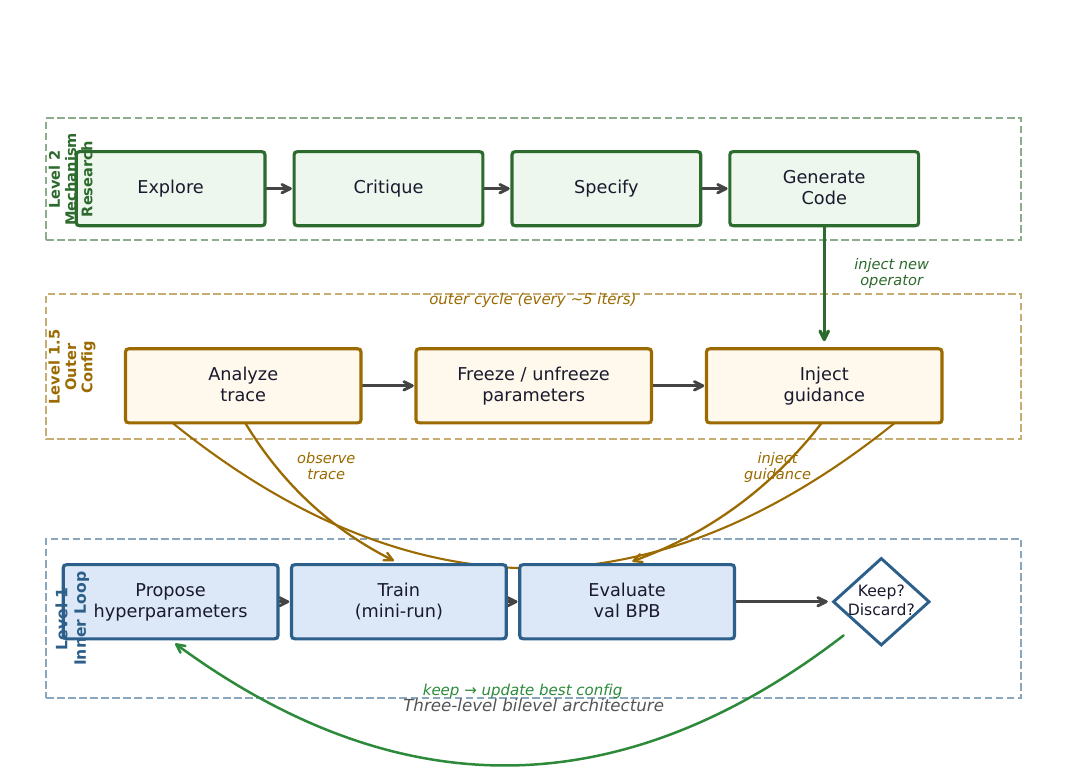}
  \caption{%
    \methodname{} architecture.
    Level~1 (blue) runs the standard propose--train--evaluate loop.
    Level~1.5 (amber) adjusts search parameters every 5 iterations.
    Level~2 (green) generates new Python mechanisms via a 4-round
    research session and injects them at runtime.%
  }
  \label{fig:architecture}
\end{figure}

\methodname{} has three nested levels (\cref{fig:architecture}):
Level~1 optimizes the task; Level~1.5 adjusts search parameters;
Level~2 generates new search \emph{mechanisms} as Python code.
All levels use the same DeepSeek \texttt{deepseek-chat} model.

\subsection{Level~1: Inner Autoresearch Loop}

The inner loop implements the standard autoresearch cycle
\citep{karpathy2026autoresearch}.
At each iteration $t$:
\begin{enumerate}
  \item The LLM receives the current \texttt{train.py} (frozen at the best
        accepted configuration), the list of active editable parameters, any
        frozen parameters, and a strategic guidance string injected by
        Level~1.5.
  \item The LLM proposes a change: a set of parameter name--value pairs and a
        one-sentence hypothesis.
  \item The change is applied to a working copy of \texttt{train.py} and
        training runs for a fixed 300-second budget.
  \item If the resulting \valbpb{} is lower than the current best, the change
        is \emph{kept} (the best copy is updated); otherwise it is
        \emph{discarded}.
\end{enumerate}
The iteration budget is fixed at 30 per repeat.
The initial configuration locks \texttt{DEPTH=8} and \texttt{ASPECT\_RATIO=64}
to prevent architecture-size changes; all other parameters (\texttt{LR},
\texttt{WEIGHT\_DECAY}, \texttt{WINDOW\_PATTERN}, \texttt{HEAD\_DIM},
\texttt{TOTAL\_BATCH\_SIZE}, etc.) are editable.

\subsection{Level~1.5: Outer Search-Strategy Loop}

Level~1.5 executes every 5 inner iterations.
It receives the full trace of proposals and outcomes and produces a
\texttt{SearchConfig} update:
\begin{itemize}
  \item \textbf{Freeze} parameters that have been proposed $\geq k$ times with
        zero net improvement (default $k=3$).
  \item \textbf{Unfreeze} parameters that were frozen but have not been
        explored since the search moved to a new region.
  \item \textbf{Inject} a guidance string instructing the inner loop to
        prioritize under-explored parameters.
\end{itemize}
Level~1.5 can redirect search diversity but cannot change the proposal
generation logic, the acceptance criterion, or the loop structure.
These structural changes require Level~2.

\subsection{Level~2: Mechanism Research and Code Injection}

Level~2 executes every 2 outer cycles.
It conducts a 4-round structured dialogue, each round making a single LLM call:

\begin{enumerate}
  \item \textbf{Explore.} The LLM reads the full \texttt{runner.py} source and
        the search trace.
        It surveys mechanisms from adjacent fields (combinatorial optimization,
        online learning, design of experiments, Bayesian optimization) and
        proposes candidate improvements.
  \item \textbf{Critique.} The LLM evaluates the candidate mechanisms against
        the observed failure mode (e.g., repetitive proposals, parameter
        fixation) and selects the most promising one.
  \item \textbf{Specify.} The LLM writes a precise interface specification:
        class name, constructor arguments, key methods with signatures, and
        integration points in \texttt{runner.py}.
  \item \textbf{Generate.} The LLM writes complete, runnable Python code
        implementing the specified mechanism, including any modifications to
        \texttt{runner.py} required to call it.
\end{enumerate}

\begin{figure}[!htbp]
  \centering
  \includegraphics[width=\columnwidth]{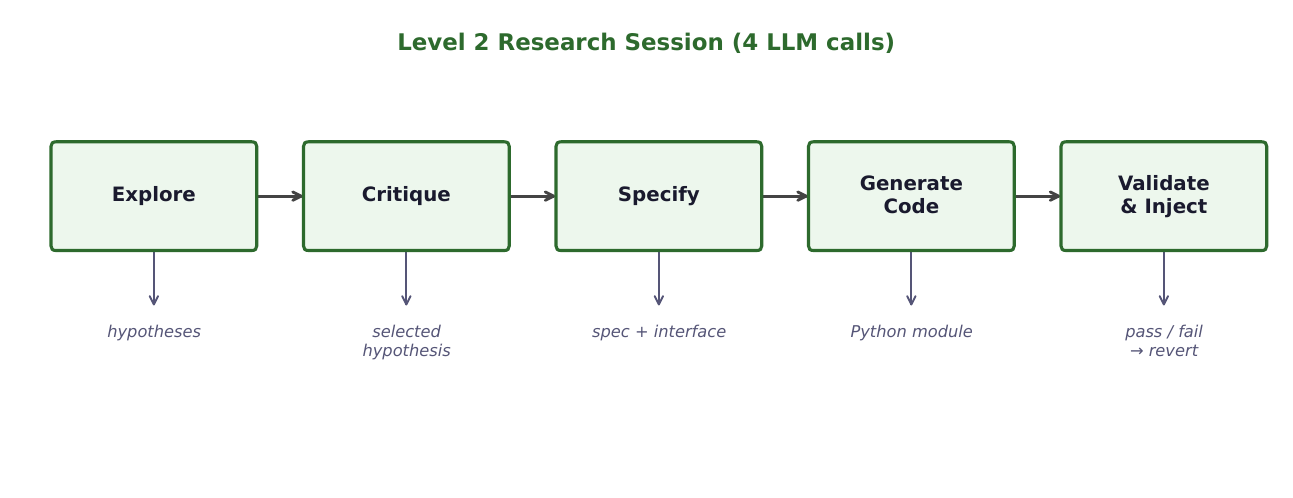}
  \caption{%
    Level 2 research session.
    Each session makes four LLM calls, producing a Python module that is
    validated before injection and, if accepted, modifies the inner loop's
    search behavior.%
  }
  \label{fig:level2session}
\end{figure}

The generated code patches \texttt{runner.py} in place and is validated
via \texttt{importlib} dynamic loading before activation.
If the import succeeds, the patched runner replaces the active one;
if it fails, the original is restored from a pre-patch backup.
This validate-and-revert mechanism prevents import-time failures from corrupting
the active runner.

\subsection{Algorithm}

\begin{algorithm}[H]
\caption{\methodname{} (Group~C configuration)}
\label{alg:bilevel}
\begin{algorithmic}
\STATE \textbf{Input:} baseline \texttt{train.py}, runner $\phi_0$, budgets
       $T{=}30$, $K{=}5$ (outer period), $M{=}2$ (L2 period)
\STATE $\theta \leftarrow$ baseline config;\; $\phi \leftarrow \phi_0$;\;
       $t \leftarrow 0$;\; outer\_cycle $\leftarrow 0$
\STATE \textsc{BestVal} $\leftarrow$ \textsc{EvaluateBaseline}($\theta$)
\STATE guidance $\leftarrow \emptyset$;\; frozen $\leftarrow \emptyset$;\;
       trace $\leftarrow \emptyset$
\WHILE{$t < T$}
  \FOR{$k = 1$ \TO $K$}
    \STATE proposal $\leftarrow$ \textsc{LLMPropose}($\theta$, $\phi$,
           guidance, frozen)
    \STATE $\theta' \leftarrow \theta \oplus$ proposal
    \STATE val $\leftarrow$ \textsc{Train}($\theta'$, budget$=$300s)
    \STATE accepted $\leftarrow$ val $<$ \textsc{BestVal}
    \IF{accepted}
      \STATE $\theta \leftarrow \theta'$;\; \textsc{BestVal} $\leftarrow$ val
    \ENDIF
    \STATE trace $\leftarrow$ trace $\cup \{(\text{proposal}, \text{val}, \text{accepted})\}$
    \STATE $t \leftarrow t + 1$
  \ENDFOR
  \STATE guidance, frozen $\leftarrow$ \textsc{Level1.5}(trace)
  \STATE outer\_cycle $\leftarrow$ outer\_cycle $+ 1$
  \IF{$t < T \;\land\; \text{outer\_cycle} \bmod M = 0$}
    \STATE $\phi' \leftarrow$ \textsc{Level2Research}($\phi$, trace)
    \IF{\textsc{ValidateImport}($\phi'$)}
      \STATE $\phi \leftarrow \phi'$
    \ELSE
      \STATE \textbf{revert} $\phi$
    \ENDIF
  \ENDIF
\ENDWHILE
\STATE \textbf{return} $\theta$
\end{algorithmic}
\end{algorithm}

\subsection{Experimental Design}

Four groups isolate the contribution of each level (\cref{tab:groups}).
All variables are held constant across groups: LLM model, GPU hardware
(RTX~5090 32\,GB, three independent servers), 300-second training budget,
30-iteration search budget, and baseline \texttt{train.py}.
Each group runs 3 independent repeats; \texttt{train.py} is restored to the
original baseline between repeats (verified by log inspection).
The primary metric is $\Delta = \text{best} - \text{baseline}$ \valbpb{}
(more negative indicates greater improvement).

\begin{table}[!htbp]
\centering
\caption{Experimental groups. All variables (LLM, GPU, budget, baseline) are held constant across groups.}
\label{tab:groups}
\begin{tabular}{@{}lll@{}}
\toprule
Group & Levels Active & Description \\
\midrule
A & Level~1 only & Pure autoresearch, no outer intervention \\
B & Level~1 + 1.5 & Inner loop plus outer strategy adjustment \\
C & Level~1 + 1.5 + 2 & Full bilevel with mechanism research \\
D & Level~1 + 2 & Inner loop plus mechanism research, no strategy adjustment \\
\bottomrule
\end{tabular}
\end{table}

%% file: sections/results.tex
\section{Results}
\label{sec:results}

\subsection{Primary Ablation Results}

\Cref{tab:main} reports \valbpb{} improvement for each group across three
independent repeats.

\begin{table}[!htbp]
\centering
\caption{%
  \valbpb{} change ($\Delta = \text{best} - \text{baseline}$, more negative
  is better) over 30 inner iterations.
  Baseline \valbpb{} varies slightly across repeats due to training
  randomness (range 1.094--1.114).
  Group~C's mean improvement is 5$\times$ that of Group~A and 7.5$\times$ that
  of Group~B (by absolute $|\Delta|$).
  Bold values indicate the best repeat within Groups~C and~D.%
}
\label{tab:main}
\begin{tabular}{@{}lrrrr@{}}
\toprule
Group & R1 & R2 & R3 & Mean $\pm$ Std \\
\midrule
A (Level~1)         & $-0.009$ & $-0.008$ & $-0.011$ & $-0.009 \pm 0.002$ \\
B (Level~1 + 1.5)   & $-0.000$ & $-0.010$ & $-0.009$ & $-0.006 \pm 0.006$ \\
C (Level~1 + 1.5 + 2) & $\mathbf{-0.065}$ & $-0.011$ & $\mathbf{-0.058}$ & $\mathbf{-0.045 \pm 0.030}$ \\
D (Level~1 + 2) & $-0.001$ & $\mathbf{-0.063}$ & $-0.039$ & $-0.034 \pm 0.031$ \\
\bottomrule
\end{tabular}
\end{table}

\begin{figure}[!htbp]
  \centering
  \includegraphics[width=\columnwidth]{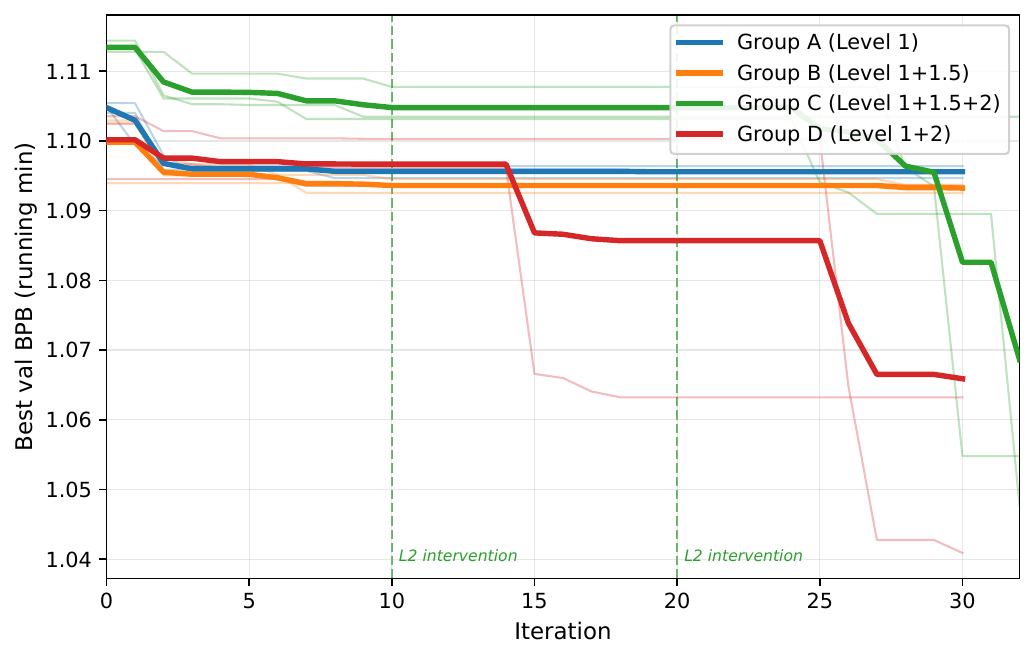}
  \caption{%
    Running-minimum \valbpb{} vs.\ iteration for all 12 runs
    (4 groups $\times$ 3 repeats).
    Thin lines: individual repeats; thick lines: group means.
    Groups~C and~D show sharp drops once their search trajectories move toward
    \texttt{TOTAL\_BATCH\_SIZE} reduction.%
  }
  \label{fig:convergence}
\end{figure}

Group~A achieves consistent but small improvements: $-0.009 \pm 0.002$.
Group~B is comparable to Group~A ($-0.006 \pm 0.006$); its R1 found
essentially no improvement ($-0.000$), inflating variance.
Group~C achieves $-0.045 \pm 0.030$, a 5$\times$ improvement over Group~A.
Two of three Group~C repeats (R1 and R3) show dramatic gains ($-0.065$ and
$-0.058$); R2 underperformed at $-0.011$.
Group~D achieves $-0.034 \pm 0.031$: R2 reached the best single result
across all D repeats ($-0.063$) while R1 barely improved ($-0.001$), giving
a mean comparable to, but slightly below, Group~C.
Group~C's separation from Groups~A and~B provides evidence that Level~2
mechanism injection is the primary driver in the full bilevel setting.
Group~D suggests that removing Level~1.5 does not necessarily prevent large
gains, but its mechanism-application failures make this ablation less clean.
\cref{fig:convergence} shows the search trajectories across all twelve runs.

\subsection{Level~2 Generated Artifacts and Mechanisms}

\Cref{tab:mechanisms} lists all code artifacts generated by Level~2 across the
six research sessions (two per repeat in Group~C), including both primary
mechanisms and auxiliary helper classes.

\begin{table}[!htbp]
\centering
\caption{%
  Level~2 generated artifacts and mechanisms.
  All code was generated on the first attempt (zero retries).
  Five of six generated code artifacts passed import validation and were
  activated; one primary mechanism (GP Regressor) was reverted due to a missing
  \texttt{sklearn}
  dependency.%
}
\label{tab:mechanisms}
\begin{tabular}{@{}llllcc@{}}
\toprule
Repeat & Round & Mechanism & Domain & Import & Active \\
\midrule
R1 & 1 & Tabu Search Manager     & Combinatorial opt.    & \checkmark & \checkmark \\
R1 & 2 & Helper class            & ---                   & \checkmark & \checkmark \\
R2 & 1 & Multi-Scale Bandit Proposer & Online learning / MAB & \checkmark & \checkmark \\
R2 & 2 & Helper class            & ---                   & \checkmark & \checkmark \\
R3 & 1 & GP Regressor            & Bayesian optimization & $\times^{*}$ & reverted \\
R3 & 2 & Syst.\ Orthogonal Exploration & DOE        & \checkmark & \checkmark \\
\bottomrule
\multicolumn{6}{@{}l}{%
  $^{*}$ Code valid but \texttt{sklearn} not installed; reverted automatically.%
}
\end{tabular}
\end{table}

The two ``Helper class'' entries are utility classes generated alongside the primary
mechanisms (e.g., data structures for tracking parameter history); they are auxiliary
to the named mechanism in each round.

\cref{fig:generated_code} shows a representative excerpt from one generated
mechanism (Tabu Search Manager, Group~C R1).
This code was written entirely by DeepSeek during a Level~2 research session;
no human edited it before injection.

\begin{figure}[!htbp]
\begin{lstlisting}[language=Python, basicstyle=\ttfamily\scriptsize,
  frame=single, numbers=left, xleftmargin=2em]
class TabuSearchManager:
  def __init__(self, tenure=5, thresholds=None):
    self.tabu_list = []
    self.tenure = tenure
    self.distance_thresholds = thresholds or {}

  def is_tabu(self, config, iteration):
    # Remove expired entries
    self.tabu_list = [e for e in self.tabu_list
                      if iteration <= e["expires_at"]]
    for entry in self.tabu_list:
      for param, tabu_val in entry["config"].items():
        if param in config:
          try:
            diff = abs(float(config[param])
                     - float(tabu_val))
            thresh = self.distance_thresholds.get(
                     param, 0.0)
            if diff <= thresh:
              return True  # too close, blocked
          except (ValueError, TypeError):
            if config[param] == tabu_val:
              return True
    return False  # allowed
\end{lstlisting}
\caption{%
  Excerpt from a Level~2 generated mechanism (Tabu Search Manager).
  This code---written entirely by the LLM during a research session---prevents
  the inner loop from revisiting recently explored parameter regions,
  breaking the deterministic proposal patterns observed in Group~A.%
}
\label{fig:generated_code}
\end{figure}

The three active named mechanisms are:
\textbf{Tabu Search Manager} (maintains a tabu list of recently visited
parameter regions, preventing the LLM from reproposing the same changes);
\textbf{Multi-Scale Bandit Proposer} (treats parameter selection as a
multi-armed bandit, balancing exploration and exploitation across parameters
at different scales); and
\textbf{Systematic Orthogonal Exploration} (forces the LLM to explore
orthogonal parameter dimensions, preventing over-focus on a single parameter).
Each active named mechanism was drawn from a different adjacent optimization
domain.
Level~2 was not told which specific mechanism to implement, though the prompt
suggested broad neighboring domains.

\subsection{Search Behavior Analysis}

The four groups exhibit qualitatively different search trajectories.

\paragraph{Group~A: near-deterministic repetition.}
All three repeats follow nearly the same proposal sequence from the same
baseline:
iteration~1 attempts \texttt{TOTAL\_BATCH\_SIZE} increase (discard);
iteration~2 reduces \texttt{WEIGHT\_DECAY} (keep, $\Delta \approx -0.008$);
iteration~3 sets \texttt{WINDOW\_PATTERN="SSSS"} (keep, $\Delta \approx -0.002$);
iterations~4--30 repeat these same two changes, accumulating up to 22
consecutive discards.
The LLM, given the same state, generates nearly the same proposals every time.

\paragraph{Group~B: redirected but bounded.}
Level~1.5 correctly identifies stalled parameters and redirects search:
by cycle~2--3 the outer loop freezes \texttt{WEIGHT\_DECAY} and
\texttt{WINDOW\_PATTERN} and redirects toward \texttt{LR},
\texttt{UNEMBEDDING\_LR}, \texttt{MATRIX\_LR}, and \texttt{FINAL\_LR\_FRAC}.
Group~B explores more parameters than Group~A, but achieves similarly sized
improvements because it operates within the same structural keep/discard
framework.

\paragraph{Group~C: Level~2 mechanisms unlock new directions.}
The decisive event in Group~C's R1 and R3 is the discovery of
\texttt{TOTAL\_BATCH\_SIZE} \emph{reduction} (from $2^{19}$ to $2^{17}$ or
$2^{18}$), which produces improvements of $-0.039$ to $-0.065$---roughly
5--8$\times$ larger than any single change found by Groups~A or~B.

\paragraph{Group~D: Level~2 without outer loop guidance.}
Group~D's pattern is more ambiguous: R2 and R3 independently discover
\texttt{TOTAL\_BATCH\_SIZE} reduction (D2 reaches $2^{17}$, D3 reaches $2^{18}$),
producing improvements of $-0.063$ and $-0.039$.
However, the per-repeat reports show that Group~D's generated mechanism patches
were not successfully applied or validated, so these repeats cannot cleanly
establish that successful mechanism injection alone caused the gains.
R1 failed to benefit: its two Level~2 sessions generated mechanisms
(\texttt{diversity\_enforcer} and \texttt{fixation\_detector}) that failed
import validation, leaving it to run as bare Level~1 with no mechanism
injection; the inner loop could not discover the batch size direction on its
own.
The absence of Level~1.5 means there is no focused parameter-freeze guidance
to steer Level~2's attention; combined with failed mechanism application in
R1, this makes Group~D outcomes more variable ($\pm 0.031$) than Group~C's
repeats, where Level~1.5 provides enriched trace context.
The Group~D mechanism-session artifacts and per-repeat reports are included in
the repository.

\subsection{The \texttt{TOTAL\_BATCH\_SIZE} Discovery}

The most impactful finding across all experiments is that reducing
\texttt{TOTAL\_BATCH\_SIZE} from $2^{19}$ to $2^{17}$--$2^{18}$ dramatically
improves \valbpb{} on the RTX~5090 under a 300-second training budget.
The mechanism is straightforward: a smaller batch size yields more gradient
steps within the fixed time budget, and better convergence for this 50M-parameter
model.
The original $2^{19}$ batch size appears better matched to the H100-oriented
setting used by the original benchmark; in our RTX~5090 setup, running SDPA
rather than Flash Attention~3, smaller batch sizes produced more gradient steps
within the fixed wall-clock budget.

Groups~A and~B both miss this direction for the same reason: DeepSeek's default
search path attempts \texttt{TOTAL\_BATCH\_SIZE} \emph{increase} first (an
implicit ``larger batch is better'' bias).
After the increase is discarded, Group~A repeats it; Group~B's outer loop
freezes \texttt{TOTAL\_BATCH\_SIZE} after the failed increase, blocking the
decrease direction entirely.
Only the high-performing Level~2-condition repeats---Group~C R1/R3 and
Group~D R2/R3---moved the search toward the decrease direction.
In Group~C, the relevant active mechanisms were Tabu Search (which prevents
revisiting failed directions) and Orthogonal Exploration (which forces
dimensional diversity).
In Group~D, the same direction appeared in the reported trajectories, but the
mechanism patches were not successfully applied or validated, so those repeats
should not be interpreted as clean evidence of successful mechanism injection.

%% file: sections/discussion.tex
\section{Discussion}
\label{sec:discussion}

\subsection{Hypothesis Testing}

The experimental design is motivated by four hypotheses.

\textbf{H1 (Group~B $>$ Group~A): Not supported.}
Group~B's mean improvement ($-0.006 \pm 0.006$) is numerically worse than
Group~A's ($-0.009 \pm 0.002$), though the difference is not meaningful given
$n=3$.
The outer loop (Level~1.5) increases search diversity---Group~B explores more
parameters than Group~A---but this diversity does not translate into larger
improvements within the 30-iteration budget.
Group~B's R1 achieved essentially zero improvement ($-0.000$), the worst
outcome of any repeat in any group.
The outer loop correctly froze stalled parameters but, having done so, the LLM
found nothing better in the remaining search space.

\textbf{H2 (Group~C $>$ Group~B): Supported.}
Group~C's mean absolute improvement ($-0.045 \pm 0.030$) is 7.5$\times$
Group~B's ($-0.006 \pm 0.006$).
Despite high variance ($\pm 0.030$), two of three repeats produced dramatic
improvements ($-0.065$, $-0.058$), and the separation between Groups~C and~B
is large relative to the within-group variance, providing meaningful evidence
that Level~2 adds value beyond Level~1.5.

\textbf{H3 (Level~2 autonomously generates and injects new mechanisms): Supported.}
Across Group~C's three independent repeats, Level~2 generated mechanisms from three
distinct active domains (combinatorial optimization, online learning, DOE)
without being told which specific mechanism to implement, though the Level~2
prompt suggested broad adjacent domains (a fourth domain, Bayesian optimization,
was attempted but reverted due to a missing dependency).
Code generation succeeded on the first attempt in all six sessions (zero
retries).
Five of six generated code artifacts passed import validation and were
activated; among the active named mechanisms, Level~2 instantiated Tabu Search,
multi-armed bandit exploration, and systematic orthogonal exploration.

\textbf{H4 (Group~D $\approx$ Group~C without Level~1.5): Suggestive but inconclusive.}
Group~D's mean ($-0.034 \pm 0.031$) is lower than Group~C's ($-0.045 \pm 0.030$),
but the difference is within the variance of both groups given $n=3$.
Group~D produced large improvements in two of three repeats, suggesting that
removing Level~1.5 does not necessarily prevent strong outcomes.
However, the per-repeat reports show that Group~D's generated mechanism patches
were not successfully applied or validated, so Group~D is not a clean test of
successful Level~2 injection without Level~1.5.
The remaining evidence suggests that Level~1.5 may provide modest robustness by
enriching the search trace that Level~2 reads, but a cleaner ablation is needed.

\subsection{Why Group~C's R2 Underperformed}

Group~C's R2 achieved only $-0.011$ improvement, comparable to Groups~A and~B,
despite receiving Level~2 mechanisms.
The most likely explanation is mechanism quality: R2's Level~2 generated the
Multi-Scale Bandit Proposer, which---while valid and correctly injected---may
be less effective than R1's Tabu Search Manager or R3's Orthogonal Exploration
at forcing exploration of the batch size dimension.
A secondary factor is computational overhead: each Level~2 research session
adds approximately 3~minutes of wall time and four additional LLM calls.
Because our experimental budget fixes the number of inner iterations rather
than total wall-clock time, this overhead does not reduce the number of inner
evaluations; it should instead be interpreted as an additional cost of the
bilevel architecture.

\subsection{Mechanism-Level Interpretation}

The main lesson is not that Python code is a privileged self-improvement
artifact.
Rather, code is a convenient carrier for a modifiable mechanism: it makes
proposal, exploration, selection, and update logic explicit enough to inspect,
generate, validate, inject, and roll back.
Other artifacts can play the same role when they change the future behavior of
an agentic loop.
A skill may encode a reusable procedural policy; a prompt may change the
agent's proposal prior; a workflow may change control flow among tools and
sub-agents; an evaluator may change which candidates survive; and a scientific
principle or world-model assumption may change the hypothesis space the agent
searches.
From this perspective, \methodname{} is best understood as mechanism-level
self-improvement: the outer loop modifies how future improvements are
generated, not merely which artifact is used in the current episode.

This also clarifies the recursive claim.
The experiments demonstrate a first bilevel step: Level~2 improves the search
behavior of Level~1.
If a discovered mechanism reliably improves Level~1, the same
mechanism-discovery principle can in principle be applied to Level~2 itself.
Such recursive bootstrapping would require additional evaluation gates,
versioning, rollback, and evidence that the meta-level process improves rather
than merely changes.
We therefore treat full recursive self-application as a natural extension, not
as an empirical result established by the present experiments.

\subsection{Limitations}

\paragraph{Small sample size.}
Three repeats per group is insufficient for rigorous statistical comparison.
Group~C's standard deviation ($\pm 0.030$) is 67\% of its absolute mean, indicating
high variability.
Reliable estimates would require $n \geq 10$ repeats per group.

\paragraph{Baseline variance.}
Baseline \valbpb{} varies across repeats (1.094--1.114) due to training
randomness from data ordering and weight initialization.
Using $\Delta = \text{best} - \text{baseline}$ mitigates this but does not
eliminate it; a lower baseline gives less headroom for improvement.
Future work should use fixed random seeds or report baseline-normalized metrics.

\paragraph{Single benchmark.}
All results are on one task: GPT pretraining at 50M parameters with a 300-second
budget on RTX~5090.
Generalization to other model sizes, training budgets, or tasks is unproven.

\paragraph{Carrier limitation.}
The experiments instantiate mechanisms as Python code.
The framework is representation-agnostic at the conceptual level, but this paper
does not systematically evaluate skills, prompts, workflows, evaluators,
scientific principles, world-model assumptions, or memory schemas as
alternative mechanism carriers.

\paragraph{Recursive limitation.}
The experiments show an outer loop improving an inner loop.
They do not demonstrate unbounded recursive self-improvement or prove that
mechanisms discovered for Level~1 will improve Level~2.
Recursive bootstrapping remains a future evaluation target.

\paragraph{Dynamic load fragility.}
A preliminary run was invalidated because the Level~2 dynamic loading pipeline
contained a \texttt{sys.modules} registration bug, causing all mechanism
injections to silently fall back to the original runner.
We fixed the bug before conducting the three reported repeats.
This episode highlights the fragility of runtime code injection: silent
fallback without error is a dangerous failure mode.

\paragraph{External dependency exposure.}
Level~2 has no constraint preventing it from importing external libraries.
One of six generated mechanisms (GP Regressor) required \texttt{sklearn}, which
was not installed.
The validate-and-revert mechanism handled this correctly, but the exposure to
arbitrary dependencies remains a reliability risk.

\paragraph{Prompt-induced domain bias.}
The Level~2 prompt explicitly suggests candidate domains (combinatorial
optimization, reinforcement learning, evolutionary algorithms, Bayesian
optimization).
This guidance is a double-edged sword: it prevents the agent from generating
irrelevant or degenerate mechanisms, but it also constrains the search space
of discoverable mechanisms to domains the prompt author anticipated.
Whether Level~2 would discover equally effective---or entirely
different---mechanisms under an unconstrained prompt remains untested.

\paragraph{No stability guarantee.}
Runtime mechanism injection can improve search behavior, but it can also
produce brittle patches, evaluator overfitting, or unsafe self-modification.
The present system relies on import validation and revert-on-failure, not on a
formal convergence or stability guarantee.

\subsection{Future Work}

Key directions include scaling to $n \geq 10$ repeats with fixed random seeds
for statistical power; evaluating on multiple benchmarks (different model
sizes, tasks, budgets) to assess generalization; testing alternative mechanism
carriers such as skills, prompts, workflows, evaluators, scientific principles,
world-model assumptions, and memory schemas;
measuring whether accepted mechanisms change the inner loop's future search
behavior across seeds and tasks; and investigating whether Level~2's code
generation quality improves with a richer interface specification or a test
harness.

%% file: sections/conclusion.tex
\section{Conclusion}
\label{sec:conclusion}

\methodname{} demonstrates that an LLM-driven outer loop can autonomously
improve an autoresearch loop by generating and injecting new search mechanisms
at runtime.
On Karpathy's GPT benchmark, Level~2 produces a 5$\times$ \valbpb{}
improvement over the inner loop alone ($-0.045$ vs.\ $-0.009$), while
parameter-level adjustment (Level~1.5) yields no reliable gain.
The generated mechanisms---drawn from combinatorial optimization, online
learning, and design of experiments---appear to improve performance by forcing
exploration of directions the LLM's default search path avoids.

Code, per-repeat reports, and mechanism-session artifacts are available at
\url{https://github.com/EdwardOptimization/Bilevel-Autoresearch}.

The core principle is empirically supported in this benchmark at the first
bilevel level:
\emph{autoresearch can improve its own search mechanism.}
This mechanism need not be limited to Python code; code is one executable
carrier among skills, prompts, workflows, evaluators, scripts, domain
principles, world-model assumptions, and memory schemas that can shape future
agent behavior.
The longer-term recursive claim is a path rather than a completed result:
mechanisms discovered for the inner loop may be fed back to improve the
meta-level loop itself, but demonstrating robust recursive self-application
requires further experiments and stronger validation gates.